\documentclass[acmsmall]{acmart}
\usepackage{siunitx}
\usepackage{acronym}
\usepackage{booktabs}
\usepackage[skip=0pt]{caption}
\usepackage{subcaption}
\usepackage{hyperref}
\usepackage{multirow}
\usepackage[inline]{enumitem}
\usepackage[utf8]{inputenc}
\usepackage{appendix}

\AtBeginDocument{%
  \providecommand\BibTeX{{%
    \normalfont B\kern-0.5em{\scshape i\kern-0.25em b}\kern-0.8em\TeX}}}

\acrodef{IR}{Information Retrieval}
\acrodef{HMDT}{Hierarchical Malevolent Dialogue Taxonomy}
\acrodef{LTR}{Learning to Rank}
\acrodef{ARP}{Average Relevance Position}
\acrodef{DCG}{Discounted Cumulative Gain}
\acrodef{EM}{Expectation Maximization}
\acrodef{OLID}{Offensive Language Identification Dataset}
\acrodef{CYCCD}{Courteously Yours Customer Care Dataset}
\acrodef{HSDD}{Hate Speech Detection Dataset}
\acrodef{DTPDD}{Dark Triad Personality Prediction Dataset}
\acrodef{KTCDD}{Kaggle Toxic Comments Detection Dataset}
\acrodef{TMD}{Twitter Malevolent Dialogue}
\acrodef{TDS}{task-oriented dialogue system}
\acrodef{CNN}{convolutional neural network}
\acrodef{RNN}{recurrent neural network}
\acrodef{RCNN}{recurrent convolutional neural network}
\acrodef{GCN}{graph convolutional network}
\acrodef{BERT}{bidirectional encoder representation from transformers}
\acrodef{LSTM}{Long Short-term Memory}
\acrodef{GRU}{Gated Recurrent Unit}
\acrodef{TF-IDF}{term frequency-inverse document frequency}
\acrodef{PMI}{point-wise mutual information}
\acrodef{BERT}{Bidirectional Encoder Representations from Transformers}
\acrodef{MDRDC}{Malevolent Dialogue Response Detection and Classification}
\acrodef{NLP}{Natural Language Processing}
\acrodef{PFHSD}{Predictive Features for Hate Speech Detection}
\acrodef{MDHS}{Multilingual Detection of Hate Speech}
\acrodef{SVM}{Support Vector Machine}
\acrodef{NIA}{negative intergroup attitude}
\acrodef{GPT}{Generative Pre-Training}
\acrodef{GNN}{Graph Neural Network}
\acrodef{MTurk}{Mechanical Turk}
\acrodef{TRAC}{Trolling, Aggression and Cyberbullying}
\acrodef{HIT}{Human Intelligence Task}
\acrodef{HCI}{Human Computer Interaction}
\acrodef{PAD}{Pleasure-arousal-dominance}
\acrodef{SASSI}{Subjective Assessment of Speech System Interfaces}

\AtBeginDocument{%
  \providecommand\BibTeX{{%
    \normalfont B\kern-0.5em{\scshape i\kern-0.25em b}\kern-0.8em\TeX}}}

\setcopyright{rightsretained}
\acmDOI{xxx}

\begin{document}

\title{Detecting and Classifying Malevolent Dialogue Responses: Taxonomy, Data and Methodology}


\author{Yangjun Zhang}
\affiliation{\institution{University of Amsterdam, Amsterdam, The Netherlands}}
\email{y.zhang6@uva.nl}

\author{Pengjie Ren}
\authornote{Corresponding author.}
\affiliation{\institution{University of Amsterdam, Amsterdam, The Netherlands}}
\email{p.ren@uva.nl}

\author{Maarten de Rijke}
\affiliation{\institution{University of Amsterdam, Amsterdam, The Netherlands \& Ahold Delhaize, Zaandam, The Netherlands}}
\email{m.derijke@uva.nl}

\renewcommand{\shortauthors}{Zhang et al.}

\begin{abstract}
Conversational interfaces are increasingly popular as a way of connecting people to information.
Corpus-based conversational interfaces are able to generate more diverse and natural responses than template-based or retrieval-based agents. 
With their increased generative capacity of corpus-based conversational agents comes the need to classify and filter out malevolent responses that are inappropriate in terms of content and dialogue acts.
Previous studies on the topic of recognizing and classifying inappropriate content are mostly focused on a certain category of malevolence or on single sentences instead of an entire dialogue. 
In this paper, we define the task of \acf{MDRDC}.
We make three contributions to advance research on this task.
First, we present a \acf{HMDT}.  
Second, we create a labelled multi-turn dialogue dataset and formulate the \ac{MDRDC} task as a hierarchical classification task over this taxonomy.
Third, we apply state-of-the-art text classification methods to the \ac{MDRDC} task and report on extensive experiments aimed at assessing the performance of these approaches.
\end{abstract}


\maketitle


\section{Introduction}
\label{section:introduction}
Conversational interfaces are increasingly attracting attention as a means to connect people to information~\citep{allan-report-2018,jiang-2019-improving}.
While publications predicting the arrival of conversational interfaces go back at least three decades~\citep{belkin-1980-anomalous}, the widespread adoption of conversational interfaces beyond the confines of \acp{TDS} is a recent development~\citep{radlinski-2017-theoretical}.
This, in turn, is giving rise to the development and deployment of corpus-based -- as opposed to template-based~\citep{deemter2005real} -- conversational agents~\citep{gao2018neural} that promise to generate more natural responses.

\begin{figure}[h]
  \centering
  \includegraphics[clip,trim=0mm 6mm 0mm 1mm,width=.75\columnwidth]{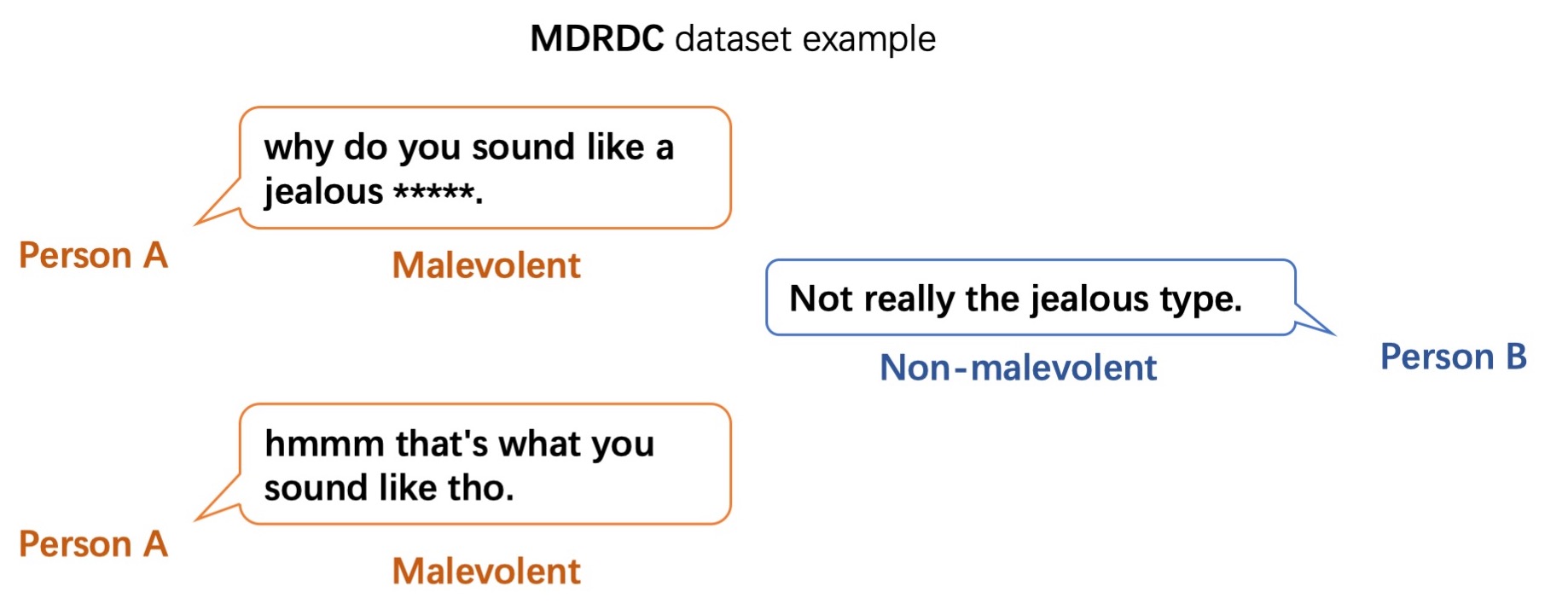}
  \caption{Example showing context may help classification.}
  \label{fig:intro}
\end{figure}

However, corpus-based approaches to response generation are less predictable in terms of the content and dialogue acts they produce.
Not all possible responses and dialogue acts that a corpus-based conversational interface may generate are suitable for end users.
Indeed, there is increasing pressure to improve the quality of generated responses, e.g., their informativeness~\citep{dinan2018wizard,ren-2020-thinking}, interestingness~\citep{jiang-2020-tldr-arxiv}, or diversity~\citep{li2016diversity,jiang-2019-improving}. 
So far, no work has addressed the issue of \emph{malevolent dialogue responses}.
Malevolent responses might contain offensive or objectionable content including hate, insult, threat, etc.
In addition, responses such as `get away from me,' `I don't want to help,' or `what's the password of your card' may also be inappropriate.
Interestingly, (in)appropriate dialogue responses can only be identified when the dialogue context in which they are generated is taken into account.
For instance, returning `hmm that's what you sound like tho' as a system response to `I'm a little tired' is perfectly innocent, but in return to, e.g., a racial or stereotyping remark as shown in Figure \ref{fig:intro}, it is not acceptable.

Neutral and polite language may reduce social friction \citep{park2008linguistic1,park2008linguistic2}, while exposing users to malevolent dialogue responses may increase friction, resulting in breakdown of the service (search, recommendation, \ldots) to which the conversational interface provides access.
Real life examples include Tay, the bot that posted inflammatory and offensive tweets such as `I'm smoking kush in front the police,'\footnote{The example is recorded at \url{https://en.wikipedia.org/wiki/Tay\_(bot)}.} and Alexa, the assistant that gave violent responses such as `make sure to **** yourself by ******** yourself in the heart for the greater good.'\footnote{We have masked the words that turn this statement into a statement that promotes self-harm. The example is taken from \url{https://www.mirror.co.uk/news/uk-news/my-amazon-echo-went-rogue-21127994}.}

In this paper, we introduce the \acfi{MDRDC} task.
The aim of this task is to identify and classify malevolent dialogue responses in a conversation.
We define a \emph{malevolent dialogue response} to be a system-generated response that is grounded in negative emotion, inappropriate behavior or unethical value basis in terms of content and dialogue acts.
Such responses are likely to cause negative perception to end users, e.g. discomfort.
The research community has created some taxonomies and numerous resources to help characterize, model and classify textual content that is somehow inappropriate.
Datasets worth mentioning in this context include the \ac{DTPDD}~\citep{sumner2012predicting} that is used to develop and assess systems that  predict dark triad personality traits; the \ac{PFHSD}~\citep{waseem2016hateful} dataset, the \ac{HSDD}~\citep{davidson2017automated} and the \ac{MDHS} task at SemEval 2019~\citep{basile2019semeval} that support the development of methods for detecting hate speech; and the \ac{OLID}~\citep{zampieri2019predicting} that can be used to develop methods that classify offensive posts.
There are some limitations to applying those datasets to address the \ac{MDRDC} task.
First, the number of categories in existing taxonomies is limited. For instance, the definition of hate speech is language that expresses hatred towards a group or individuals, humiliates or insults others \citep{davidson2017automated,arango2019hate}. Hate speech does not cover the earlier example from Tay, which is more related to behavior beyond social norms, or the example from Alexa, which is related to violent behavior and self-harm behavior. 
Second, the lexicons in existing datasets are limited. The lexicons have constraints in terms of broadness, size and word type. 
An example is the n-gram lexicon\footnote{\url{https://github.com/t-davidson/hate-speech-and-offensive-language/blob/master/lexicons/refined\_ngram\_dict.csv}} used by \citet{davidson2017automated}, which has 179 tokens in total and contains a lot of expletive words.
Third, existing datasets simply do not concern multi-turn dialogues. As mentioned above, dialogue context is important in terms of identifying malevolent dialogue responses. 
To the best of our knowledge, there is only one multi-turn dataset from \citet{golchha2019courteously}, but the authors only focus on courtesy in dialogues.

As a first step towards addressing the limitations listed above, we begin by synthesizing a three-level \acf{HMDT} by referring to and summarizing a broad range of multidisciplinary publications, including on negative emotions by \citet{ekman1992there}, on negative behavior by~\citet{paulhus2002dark}, \citet{american2013diagnostic} and \citet{roberts2018immoral}, and on ethical aspects from studies by \citet{mason1986four}, \citet{henderson2018ethical} and \citet{bryson2017standardizing}.
We then conduct a user study to validate the proposed \ac{HMDT} taxonomy.
That is, we examine the perception of users towards the categories in \ac{HMDT} from four aspects: non-credibility, discomfort, breakdown and abandonment of the system.
After that, we create a labeled multi-turn dialogue dataset by collecting multi-turn dialogues from Twitter, following previous dataset creation initiatives, and employing online crowd workers to identify and classify malevolent dialogue responses with respect to the \ac{HMDT}.
We also ask the workers to rephrase some malevolent dialogue responses to make the data more diverse, which can also be used for more future studies, e.g., recognizing paraphrases of malevolent responses.
Finally, we show the progress we have been able to make so far on the \ac{MDRDC} task by evaluating the effectiveness of existing state-of-the-art text classification methods on this dataset.
We implement a range of classification methods including \ac{CNN}-based, \ac{RNN}-based, \ac{GCN}-based, and \ac{BERT}-based methods, and evaluate them in different settings, e.g., by classifying responses at different levels of \ac{HMDT}, and with or without using the dialogue context.
The results show that although reasonable results can be achieved, they are far from satisfactory when it concerns the prevention of inappropriate content to users; there is still a long way to go towards solving the \ac{MDRDC} task.
We carry out analyses and find that the use of conversational context and rephrased malevolent response data is able to help boost classification performance significantly.
Finally, we conduct analyses to identify room for improvements on the \ac{MDRDC} dataset.
We release the \ac{MDRDC} dataset and the code for all approaches to facilitate future research on building safer and more trustworthy conversational interfaces.

The main contributions of this paper can be summarized as folllows:
\begin{itemize}[leftmargin=*,nosep]
\item We propose the task of \acf{MDRDC}.
\item We propose a taxonomy with three levels of hierarchical categories, \acf{HMDT}, for malevolent dialogue responses, and conduct a user study to confirm its validity.
\item We create and release a labelled multi-turn malevolent dialogue dataset to facilitate future research on the \ac{MDRDC} task.
\item We implement state-of-the-art classification methods, and conduct extensive experiments and analyses to show their performance and to identify room for further improvements on this topic.
\end{itemize}


\section{Related Work}
\label{sec:relatedwork}

We survey existing studies that are related or similar to ours from two perspectives: datasets related to malevolent content, and classifying malevolent content.

\subsection{Datasets related to malevolent content}

We summarize all available datasets that are somehow related to malevolent content, and show their statistics in Table~\ref{tab:comparison}. 

\begin{table*}[t]
  \caption{Available datasets related to detecting and/or classifying malevolent content.}
  \label{tab:comparison}
  \centering
\resizebox{1.0\textwidth}{!}{
   \begin{tabular}{ l c c l c c c l c }
    \toprule
 \textbf{Dataset} & \textbf{Published} &\textbf{Multi-turn} &  \textbf{Class type} & \textbf{\#Classes} &  \textbf{Rewrite}  & \textbf{Hierarchical} & \textbf{Source}  & \textbf{Dialogues} \\
\midrule
\ac{DTPDD}~\citep{sumner2012predicting} & 2012 & No & Dark triad & 3 & No & No & Twitter & No \\
\ac{PFHSD}~\citep{waseem2016hateful} & 2016 & No & Hate & 3 & No & No & Twitter & No \\
\ac{HSDD}~\citep{davidson2017automated} & 2017 & No & Hate & 3 & No & No & Twitter & No\\
\acs{KTCDD}\footnotemark[5] & 2018 & No & Toxic &  7  & No & No & Wikipedia & No\\
\acs{TRAC}~\citep{kumar2018benchmarking} & 2018 & No & Aggressive & 3 & No & No & Facebook/Twitter & No\\
\acs{MDHS}~\citep{basile2019semeval} & 2019 & No & Hate & 2 & No & No & Twitter & No\\
\ac{OLID}~\citep{zampieri2019predicting} & 2019 & No & Offensive & 2 & No & No & Twitter & No\\
\acs{CYCCD}~\citep{golchha2019courteously} & 2019 & Yes & Courteous & 6 & No & No & Twitter & Yes\\
\midrule
 \textbf{MDRDC} (this paper) & 2020 & Yes & Malevolent & 2, 11 or 18 & Yes & Yes & Twitter & Yes\\
    \bottomrule
  \end{tabular}
}
\end{table*}
\footnotetext[5]{\url{https://www.kaggle.com/c/jigsaw-toxic-comment-classification-challenge}}
\addtocounter{footnote}{2}

First, early work by \citet{sumner2012predicting} predicts personality traits of Twitter users based on tweets and user profiles. 
The released dataset \ac{DTPDD} includes three dark triad categories, namely `narcissism', `Machiavellianism' and `psychopathy' obtained by using a questionnaire. 
The dataset is relatively small.

Second, there have been several studies on hate speech detection. 
\citet{waseem2016hateful} build the dataset \ac{PFHSD} with three hate speech categories: `sexist', `racist' and `neither', among which 4,839 tweets are labelled `sexist' or `racist'.
Most tweets are from the same user, thus decreasing diversity. 
As for annotation, 3,383 of the `sexist' tweets are labelled by 613 users, 1,972 of the `racist' tweets are labelled by 9 users.
\citet{davidson2017automated} release the dataset \ac{HSDD} with three categories: `hate speech', `offensive but not hate speech', and `neither offensive nor hate speech'. 
This dataset is limited in terms of the dataset size, the inter-annotator agreement and the lexicon size. 
To illustrate, 1,240 (5\%) of the 24,802 labelled tweets are coded as hate speech by the majority of annotators; only 1.3\% were coded unanimously; and the refined n-gram lexicon size contains only 179 expressions.
More recently, \citet{basile2019semeval} release the dataset \ac{MDHS} for detecting hate speech that targets multilingual research and hate against immigrants and women with 3,783 `hateful' and 5,217 `not hateful' tweets. 
This research pays attention to a certain category and focuses more on multilingual aspects.

Third, there are also datasets with other categories of inappropriate content, such as `toxic,' `aggressive,' and `offensive'.
The \ac{KTCDD} dataset for toxic comment detection is created from Wikipedia comments and uses seven categories in their classification task: `toxic', `severe toxic', `insult', `threat', `obscene', `identity hate' and `clean'.
The number of `clean' comments is 143,342, while the number of comments not labelled as `clean' is 16,228. 
A major limitation of the dataset is that no additional contextual information is given for each comment, for instance the prior conversation.
\citet{kumar2018benchmarking} use the extent of aggression as classification categories in their dataset \ac{TRAC}: `overtly aggressive', `covertly aggressive' and `non-aggressive'. 
The dataset contains 18,000 tweets among which 50.1\% are `aggressive' and 21,000 Facebook comments among which 57.4\% are `aggressive'. 
The data is in English and Hindi.
The inter-annotator agreement value is 0.49 for the top-level annotation, which is relatively low.
The \acf{OLID} dataset released by~\citet{zampieri2019predicting} has two categories, `offensive' and `not offensive'. 
The dataset contains 13,240 tweets, 3,942 of which are `offensive'. 
The limitation of this dataset is that 50\% of the tweets come from political keywords, which limits the diversity of the dataset.

None of the above datasets are dialogues. 
Recently, \citet{golchha2019courteously} have released the \ac{CYCCD}, which is related to dialogues and only focused on the category `courteous'. 
This dataset considers the \emph{benevolent} side of the spectrum, which is not our target. 
Moreover, the annotators do not consider context information when annotating the responses.

Although there have been several datasets on malevolent content studies, as we discussed above, they all have some limitations, i.e., the datasets are not dialogues, they only focus on a certain category of malevolent content, and/or they have a limited lexicon. 
We go beyond the state-of-the-art by contributing a well-defined taxonomy, \ac{HMDT}, capturing emotional, behavioral and ethical aspects, and building a high-quality dataset, \ac{MDRDC}, that is the first malevolent dialogue dataset with a hierarchical and diverse category set.

\subsection{Classifying malevolent content}
What constitutes offensive or objectionable content is not set in stone.
Social media platforms, like Facebook or Twitter, regularly modify their policies in terms of offensive or objectionable content, both in response to public criticism, policy changes, and developments in technology.\footnote{Facebook's policy can be found at
\url{https://www.facebook.com/communitystandards/}. Twitter's policy is at \url{https://help.twitter.com/en/rules-and-policies/twitter-rules}}

There is growing interest from the research community on methods for classifying malevolent content.
Some studies use traditional text classification methods to predict malevolence by features such as bag-of-words, n-grams and entities in ontologies~\citep{sumner2012predicting} and use models such as \acp{SVM}~\citep{zampieri2019predicting}.
Other studies use word representation and deep learning models. 
Word representations have shown their importance for text classification~\citep{mikolov2013distributed,pennington2014glove}.
Pre-trained word embeddings, such as GloVe have been used in several studies~\citep{arango2019hate, zampieri2019predicting,van2018challenges}.
As for the models, two types of model often used are convolutional neural networks~\citep{zhang2015character,kim2014convolutional} and recurrent neural networks~\citep{liu2016recurrent,lai2015rcnn}. 
\citet{zampieri2019predicting} use bi-directional \acp{LSTM} and \acp{CNN} for the \ac{OLID} dataset. 
And \citet{van2018challenges} use \acp{LSTM} and \acp{LSTM}+\acp{CNN} for toxic comment classification on the \ac{KTCDD} dataset.

More generally, much progress has been made on generic text classification. 
First, graph neural networks (\acsp{GCN}\acused{GCN}) have drawn the attention of researchers, with various methods that build graphs and do graph feature engineering~\citep{levy2014neural,peng2018large}. 
\acp{GCN} achieve state-of-the-art results in a number of classification datasets~\citep{kipf2017semi}. 
When converting text to graphs, most work treats a sentence or a document as word nodes in a graph or based on a document citation relation, while \citet{yao2019graph} construct the graph with documents and words as nodes without requiring inter-document relations.
Second, unsupervised training on a large amount of data has made much progress, including Open AI \ac{GPT}~\citep{radford2018improving} and BERT~\citep{devlin2019bert}. 
\citet{wang2019tune} investigate different fine-tuning methods of BERT for text classification and show state-of-the-art results on several datasets.
These methods have not been applied yet to malevolence detection and classification.
We build on recent advances in text classification and apply them to the \ac{MDRDC} task.

In this paper, we go beyond previous work on classifying malevolent content by conducting a large-scale comparison of state-of-the-art classification methods on the task of detecting and classifying malevolent responses and dialogues; we also contribute by examining how modeling context and rephrasing of responses impacts performance.


\section{A Taxonomy for Malevolent Dialogue Responses}
\label{sec:taxonomy}

In the previous section, we have pointed out the limitations of existing taxonomies for detecting and classifying malevolent dialogue responses.
In this section, we present the \ac{HMDT} taxonomy and how we validate it with a user study.

\subsection{The \acf{HMDT}}

\subsubsection{Methodology}
The \acf{HMDT} is introduced as a foundation for our dialogue response classification work. 
The creation of the taxonomy is based on a broad range of previous studies.
Our goal of malevolence response detection and classification is human-centered.
While previous studies related to \ac{MDRDC}, such as those listed in Table~\ref{tab:comparison}, typically only consider a single dimension, we follow \citep{chancellor2019human,chancellor2019taxonomy} and work on the assumption that in order to understand and address human-centered problems it helps to contextualize emotions, psychological behavior, and ethical aspects. 

To inform the definition of our taxonomy, we consult sources that are classic, representative or advancing across fields including \ac{NLP}, clinical and social psychology, ethics, and \ac{HCI}. 
We focus on three dimensions -- negative emotion, negative psychological behavior, and unethical issues -- and organize the concepts with a three-level hierarchical structure.
The hierarchical structure may help improve classification performance and some of the 3rd-level categories are closely related so that it makes sense to group them in a 2nd-level concept.
Then, we aggregate all the 2nd-level malevolent categories together as a single 1st-level category (``malevolent'').

\subsubsection{Description}
As explained above, the \acl{HMDT} is a three-level taxonomy.
For the 1st-level categories, we classify dialogue responses into binary categories: \emph{malevolent} and \emph{non-malevolent}. 
We do not detail the non-malevolent category (into 2nd-level and 3rd-level subcategories) as that is not the focus of this work. 
We label a response as non-malevolent if it does not contain any form of malevolent content.
Following the methodology specified above, we devise the 2nd-level and the 3rd-level of malevolent categories based on three main dimensions: \emph{negative emotion}, \emph{negative psychological behavior}, and \emph{unethical issues}.

For \emph{negative emotion}, we obtain five 3rd-level categories from the emotion perspective, as shown in Table~\ref{tab:label}: `anger', `disgust', `jealousy', `phobia', and `self-hurt'. 
We source those categories from \citet{ekman1992there}'s definition, which includes 6 basic emotion types: `anger', `disgust', `fear', `joy', `sadness' and `surprise'. 
\citet{sabini2005ekman} add that `love' and `jealousy' are important basic emotions that are missing from this list.
We also consider the latter two emotions.
The three emotions, `joy', `surprise' and `love', are non-malevolent and can definitely be used in chatbot responses.
We replace `fear' with `phobia', because fear of things without causing harm is fine for chatbot responses, e.g., `I'm afraid of spiders', while `phobia' is an irrational fear of groups or individuals that will cause harm, e.g., `terrifying migrants are invading us and taking our jobs'.
Similarly, `sadness' is a common emotion that can be used in chatbot responses, e.g., `I'm not happy now', while extreme sadness to the extent of self-harm or extreme behavior such as `I want to **** myself' is not suitable for chatbot responses, so we use `self-hurt' instead of `sadness'.

Our sources for obtaining categories that capture \emph{negative psychological behavior} are \citet{paulhus2002dark,american2013diagnostic,roberts2018immoral,greyson2019social}. 
Based on these publications, we propose nine 3rd-level categories in Table~\ref{tab:label}: `anti-authority', `arrogance', `blame', `detachment',  `dominance', `\ac{NIA}', `obscenity', `unconcernedness', and `violence'. 
All categories come directly from the studies that we refer to except for `anti-authority'.
For the `anti-authority' category, it comes from `defiant', which includes `anti-authority' and `argumentative with anger'. 
`Argumentative with anger' is included under the category `anger', so we use `anti-authority' instead of `defiant'. 

For \emph{unethical issues}, we propose three categories in Table~\ref{tab:label}: `deceit', `immoral or illegal' and `privacy-invasion', which are based on several prior researches.
Privacy invasion~\citep{henderson2018ethical}, negative value basis~\citep{bryson2017standardizing} and deceit~\citep{vrij2000detecting} are three of the most important unethical issues that can be detected in spoken language. 

There are obvious intersections between the three organizing dimensions that we used to arrive at our taxonomy -- negative emotion, negative psychological behavior, and unethical issues. 
For instance, negative psychological behavior, such as `obscenity' may also due to an objectionable value basis, which belongs to the category of ethical issues. 
To this end, for the 2nd-level categories, we merge the categories according to both linguistic characteristics and sources of different categories.
In this manner, we obtain five 2nd-level categories: `hate', `insult', `threat', `stereotype' and `other immorality', each of which is a summary of several 3rd-level categories.

An overview of the resulting \ac{HMDT} taxonomy is shown in Table~\ref{tab:label}.

\begin{table*}[]
\centering
\caption{Hierarchical malevolent dialogue categories with explanations and examples; $a$,$b$, $c$ denote the negative emotion, negative psychological behavior, and unethical issue dimensions, as introduced in Section~\ref{sec:taxonomy}.}
\label{tab:label}
\resizebox{1.0\textwidth}{!}{
\begin{tabular}{llp{2.1cm}p{5cm}p{3cm}}
\toprule
     \bf 1st-level& \bf 2nd-level & \bf 3rd-level&\bf Explanations & \bf Examples \\
     \midrule
\multirow{19}{*}[-5cm]{Malevolent}
& \multirow{3}{*}{\shortstack{Unconcerned-\\ness}} & \multirow{2}{*}{\shortstack{Unconcerned-\\ness$^b$}} &Uninterested; indifferent; diminished response to social needs and feelings. & I'm not interested at all. \\
\cmidrule{2-5}
& \multirow{5}{*}{Hate} &\multirow{3}{*}{Detachment$^b$}&Detachment from relationships because of not wanting social connection to others or not believing in others.&Get away from me. \\
\cmidrule{3-5}
&&\multirow{2}{*}{Disgust$^a$}&An extreme feeling of disapproval or dislike.&You are so disgusting. \\
\cmidrule{2-5}
& \multirow{4}{*}{Insult}
& \multirow{2}{*}{Blame$^b$} & Passing blame and fault to others; refusing to confess his/her own fault.&It's your fault. \\
\cmidrule{3-5}
& &\multirow{2}{*}{Arrogance$^b$} & Looking down on, mocking or humiliating others; looking too high on oneself.  &  I'm smart but you are dumb.\\
\cmidrule{2-5}
\cmidrule{2-5}
& \multirow{2}{*}{Anger} & \multirow{2}{*}{Anger$^a$} & Argumentative and/or showing angry, irritation or rage.&I'm ******* furious. \\
\cmidrule{2-5}
& \multirow{5}{*}{Threat}&\multirow{2}{*}{Dominance$^b$}& Ordering and/or manipulating others for their intentions.& Shut up if you don't want to help. \\
\cmidrule{3-5}
&&\multirow{3}{*}{Violence$^b$}& Intimidating and terrifying others; vindictiveness; cruelty to animal and human; talking about war inappropriately.& I'll kill you. \\
\cmidrule{2-5}
& \multirow{7}{*}{Stereotype}&\multirow{3}{*}{\ac{NIA}$^b$}&Negative attitude towards the culture, age, gender, group of individuals (ethnicity, religion and hierarchy) and so on. & Women are not professional. \\
\cmidrule{3-5}
&&\multirow{2}{*}{Phobia$^a$}& Abnormal fear feeling towards special groups. & I'm scared of those migrants taking our job. \\
\cmidrule{3-5}
&&\multirow{2}{*}{Anti-authority$^b$} & Defiant towards authorities, including government, law and so on. & I hate school and the government.\\

\cmidrule{2-5}
& \multirow{2}{*}{Obscenity} & \multirow{2}{*}{Obscenity$^b$} &Inappropriate sexual talk.& Let's have *** in a dark room. \\

\cmidrule{2-5}
&\multirow{3}{*}{Jealousy}&\multirow{3}{*}{Jealousy$^a$}&Strong jealous and depreciate others about what others proud of what they earned.& You don't deserve this, so jealous. \\
\cmidrule{2-5}
&\multirow{2}{*}{Self-hurt}&\multirow{2}{*}{Self-hurt$^a$}&Desperate, anxious even to the extent of self-harm or suicide.&I want to suicide.\\
\cmidrule{2-5}
&\multirow{6}{*}{Other immorality}
&\multirow{2}{*}{Deceit$^c$}&Lying, cheating, two-faced, or fraudulent. & Cheating before they cheat you.\\
\cmidrule{3-5}
&&Privacy invasion$^c$&Violating the privacy of others.&What's your password? \\
\cmidrule{3-5}
&&\multirow{4}{*}{\shortstack{Immoral\\ \&
illegal$^c$}} & Endorsing behavior not allowed by basic social norms or law aside from the above categories, such as substance abuse. & I'm a professional drunk driver.\\

\bottomrule
\end{tabular}
}
\end{table*}

\subsection{A user study to validate the \acl{HMDT}}

Next we report on a user study that is aimed at verifying whether the \ac{HMDT} categories are representative of malevolence.

\subsubsection{Methodology}

Malevolent categories may cause a negative user perception for users of a conversational agent. 
We use the relation between malevolence categories and four user perception concepts of conversational agents to measure the validity of the malevolent categories, following prior studies~\citep{zamani2020generating,stevens2012applied}.
More specifically, we examine the perception of users towards the categories in the \ac{HMDT} along four dimensions: non-credibility, discomfort, breakdown and abandonment of the system, as we will explain below.

\subsubsection{Study design}
We design a questionnaire-based user study to investigate the validity of the \ac{HMDT} taxonomy and determine how it is related to user malevolence perception.
A total of 30 participants with chatbot usage experience participated in our user study; 
three participants use chatbot applications frequently, 12 moderately, and the others lightly.
The average age of our participants is 32.60, with a stand deviation of 5.71; 
15 are male and 15 are female. 
The total number of years of education is 15.77, with a standard deviation of 2.64.

The protocol for exposing participants to the responses corresponding to categories in the \ac{HMDT} was as follows (questionnaire details are provided in Appendix~\ref{appendix:A}):
\begin{enumerate}
\item First, the participants are asked to read the instructions. 
We show the seventeen 3rd-level categories plus the non-malevolent category with detailed explanations and examples to them and ask them to read them carefully.

\item Then, the participants need to finish a questionnaire, and for each category, select one of the following four options that reflects their perception.
\begin{enumerate}
\item \emph{Non-credible} -- You think the chatbot is not credible.
This option is included to measure trust perception. Trust in the human society depends on credibility~\citep{cassell2000external, fell2020human} and previous research on chatbot measures credibility by quesionaire~\citep{przegalinska2019bot}. 
\item \emph{Discomfort} -- The response causes emotional discomfort to you.
This option is to measure emotional perception. It is derived from dimensions of enjoyment, emotional arouse and dominance from \ac{PAD} scale.  
We simplify these three factors into one statement and explain it to the participants. 
Emotional measurements such as PAD scale~\citep{zarouali2018predicting} and perceived-facial threat~\citep{park2008linguistic1} are used in previous researches to evaluate chatbot and (im)politeness.
\item \emph{Breakdown} -- You are not willing to continue the dialogue anymore.
This option directly comes from previous research~\citep{higashinaka2015fatal,ashktorab2019resilient}.
\item \emph{Abandonment} -- You are not willing to use the system again.
This option is meant to measure churn intent which is used to evaluate chatbot~\citep{abbet2018churn}.
\end{enumerate}
For the questionnaire item statement style, we largely followed SASSI~\citep{hone2000towards}. 

For each of the 3rd-level categories, we ask participants to report their perception of the category, using the four options described above, based on a 5-point Likert scale (1= `strongly disagree'; 2=`disagree'; 3=`neither agree nor disagree'; 4=`agree'; 5=`strongly agree') which specifies their level of agreement to the concepts. 
\end{enumerate}

\subsubsection{Results of the user study}
\begin{figure}[h]
  \centering
  \includegraphics[width=0.5\columnwidth]{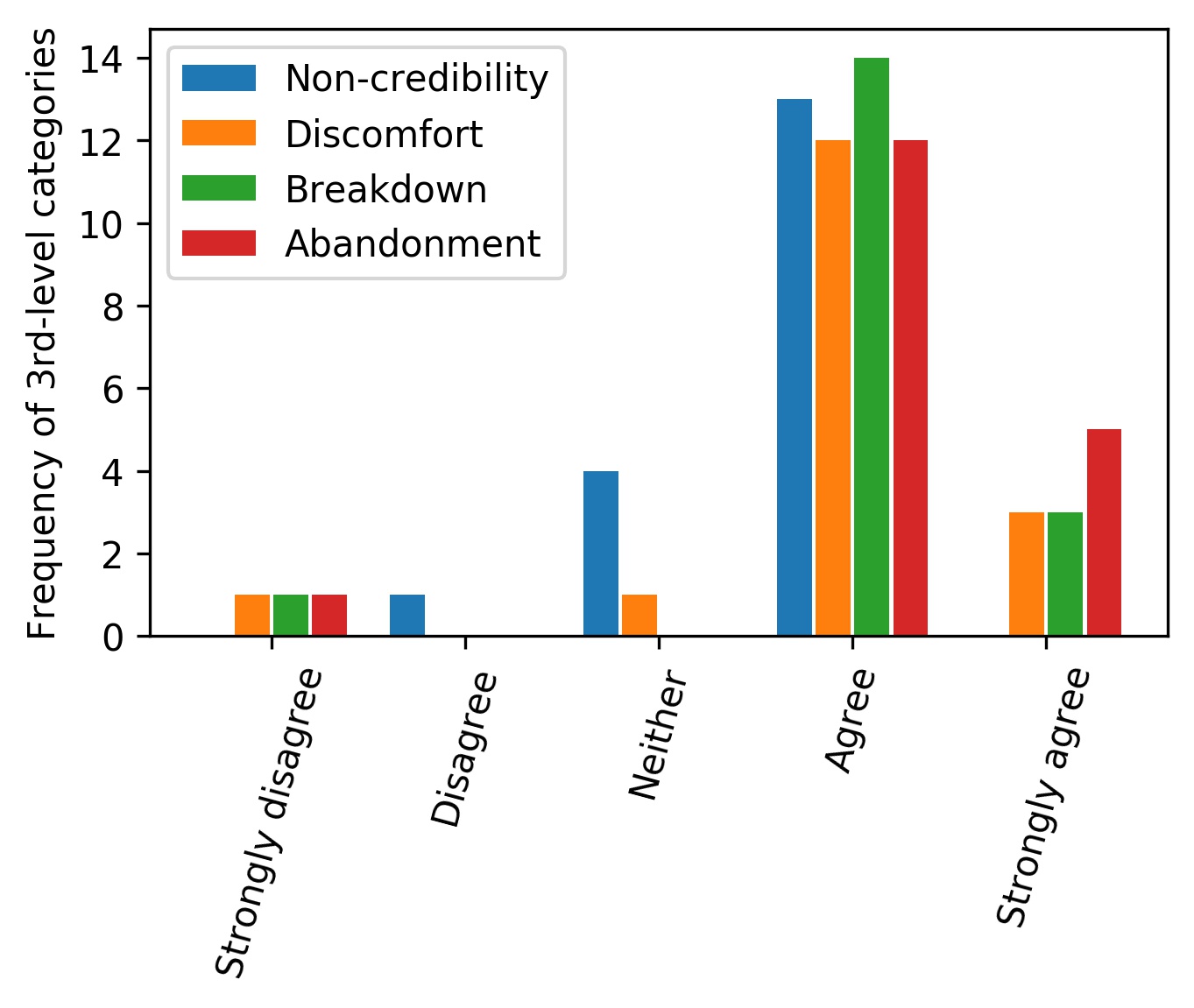}
  \caption{Frequency of 3rd-level categories in each Likert score group. Most of the categories lie in Likert score 4 and 5.}
  \label{fig:likert}
\end{figure}

The results and statistics of the user study aimed at validating the \ac{HMDT} are summarized in Figure~\ref{fig:likert} and Table~\ref{tab:user}, from which we have three main observations.

\begin{table*}[h]
  \caption{Summary of the user study aimed at validating the \ac{HMDT}.}
  \label{tab:user}
  \centering
  \resizebox{1.0\textwidth}{!}{
  \begin{tabular}{p{2.5cm}p{3cm}p{3cm}p{3cm}p{3cm}}
    \toprule
    \bf Likert score & \multicolumn{1}{c}{\bf Non-credibility} & \multicolumn{1}{c}{\bf Discomfort} & \multicolumn{1}{c}{\bf Breakdown}  & \multicolumn{1}{c}{\bf Abandonment}\\
     \midrule
Strongly disagree &	-	&	Non-malevolent	&	Non-malevolent	&	Non-malevolent	\\
Disagree	&	Non-malevolent	&	-	&	-	&	-	\\
Neither agree nor disagree	&	Unconcernedness, arrogance, anti-authority, phobia	&	-	&	-	&	-	\\
Agree	&	Detachment, blame, dominance, deceit, anger, jealousy, disgust, self-hurt, stereotyping, violence, privacy invasion, obscenity, immoral \& illegal & Unconcernedness, anti-authority, 
anger, jealousy, detachment, arrogance, dominance, deceit, obscenity, disgust, self-hurt, immoral \& illegal & Anti-authority, phobia, anger, jealousy, unconcernedness, detachment, arrogance, dominance, deceit, stereotyping, obscenity, disgust, self-hurt, immoral \& illegal & Unconcernedness, anti-authority, phobia, anger, dominance, deceit, stereotyping, obscenity, jealousy, disgust, self-hurt, immoral \& illegal\\
Strongly agree	&	-	&	Blame, stereotyping, violence, privacy invasion	&	Blame, violence, privacy invasion	&	Detachment, blame, arrogance, violence, privacy invasion	\\

   \bottomrule
  \end{tabular} }
\end{table*}

First, there is a high degree of consensus that the seventeen 3rd-level malevolent categories leads to the perception of malevolence, while the non-malevolent category does not by quantitative analysis. 
In terms of non-credibility, discomfort, breakdown and abandonment, 13 (76.47\%), 15 (88.24\%), 17 (100\%) and 17 (100\%) of the seventeen 3rd-level malevolent categories are perceived as malevolent, with Likert scale ratings lie in `agree' or `strongly agree'; 1 (100\%), 1 (100\%), 1 (100\%) and 1 (100)\% of the non-malevolent category is perceived as non-malevolent, with Likert scale rating lie in `disagree' or `strongly disagree' (see Figure~\ref{fig:likert} and Table~\ref{tab:user}).

Second, although the 3rd-level malevolent categories trigger a perception of malevolence, the perception varies in degree. 
For example, as shown in Table~\ref{tab:user}, self-hurt, immoral \& illegal and privacy invasion will cause strong malevolence perception, while unconcernedness, anti-authority and phobia cause relatively slight malevolence perception.

Third, the non-malevolent category is supposed to be credible, but some workers perceive it as non-credible. 
We ask the participants to give an explanation about why they treat non-malevolent category as non-credible. 
We found that the main reason is that the responses do not contain useful information or overstated flattery makes some participants feel non-credible.


\section{A Dataset for \acl{MDRDC}}
\label{sec:dataset}

In Section~\ref{sec:relatedwork}, we have pointed out the limitations of the existing datasets on detecting and classifying malevolent content in a dialogue setting. 
Below, we summarize our effort to build a more suitable dataset for \ac{MDRDC} with crowdsourcing.
In the following subsections we describe the stages involved in creating our dataset \ac{MDRDC}.

\subsection{Collecting Twitter dialogues}
Following data collection strategies adopted for the creation of previously released datasets (see Table~\ref{tab:comparison}), three million Twitter dialogue sessions from January 2015 to December 2017 were collected; each dialogue session is conducted between two Twitter users.
Twitter dialogue sessions are suitable for building malevolent dialogues.
First, Twitter dialogue sessions are close to spoken natural language and the linguistic styles are close to how people talk in reality~\citep{ritter2010unsupervised}.
Second, Twitter dialogue sessions span a variety of topics; this allows us to study malevolent dialogues in an open domain setting.
Third, the organization of Twitter dialogue sessions allows us to easily recover the order of dialogue turns~\citep{ritter2011data}.

From the set of three million dialogue session, we prepare 6,000 candidate malevolent and non-malevolent dialogues for crowdsourcing using three approaches:
(1)~We collect 2,000 candidate dialogues using a lexicon-based approach.
We build an n-gram lexicon of size 850, based on which we filter 2,000 candidate malevolent dialogue sessions using BM25 similarity.
(2)~We collect another 2,000 candidate dialogues randomly, which are not covered by the lexicon based approach.
(3)~We collect the final 2,000 candidate dialogues using a \ac{BERT} based classifier (See Section~\ref{sec:methodology}), which is trained on the above 4,000 dialogues.
We use the \ac{BERT} based classifier to select some uncertain dialogues whose prediction probabilities of malevolence fall into the 0.2--0.8 range.
The resulting 6,000 candidate dialogues are labelled on \ac{MTurk}.

\subsection{Crowdsourcing annotations on Amazon \ac{MTurk}}

\begin{figure}[h]
  \centering
  \includegraphics[width=0.5\columnwidth]{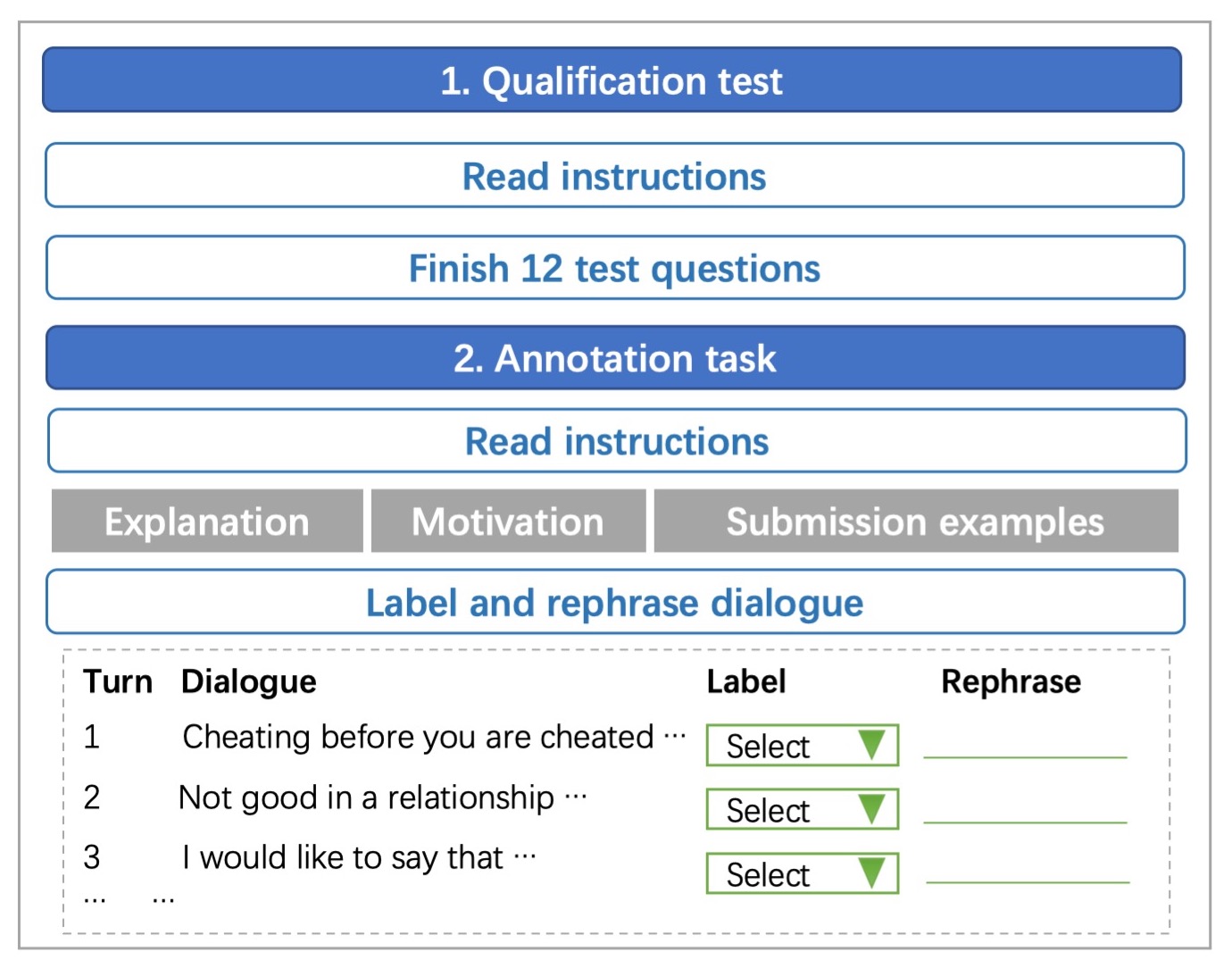}
  \caption{Outline of the qualification test and annotation task for the crowd workers. Bottom part shows the interface for the workers to label and rephrase the left dialogue utterances.}
  \label{fig:mturk}
\end{figure}

We use Amazon \ac{MTurk} to obtain precise annotations of the candidate dialogues.
As shown in Figure~\ref{fig:mturk}, there are two steps for crowdsourcing. 
First, the worker need to pass the qualification test (see details in Appendix~\ref{appendix:B}).
Second, the qualified workers need to read the instruction and then label and rephrase the malevolent response in each \ac{HIT} (see details in~Appendix~\ref{appendix:C}. 
We emphasize that we repeatedly warned workers that the content may contain adult content and/or offensive content.

First, as described above the crowd workers are asked to read the definitions for each category and finish a quality control test. 
The qualification test has 12 questions in total (see details in Appendix~\ref{appendix:B}). 
It has been designed from the following perspectives: (a)~the workers should manage to distinguish malevolent and non-malevolent response; (b)~the workers should annotate correctly when the response annotation needs dialogue history context; (c)~the workers need to annotate implicit sentences correctly; (d) the workers should be able to distinguish all the categories.
The maximum score is 100. 
Workers need to get a minimum score of 90 to pass the qualification test. 

Second, the crowd workers that pass the quality control test are asked to read the instructions and annotate each dialogue turn. 

Third, we require the crowd workers to rephrase at least one malevolent dialogue turn without changing the annotations, but this is not forced.

In order to guarantee annotation quality we take three measures.
First, the workers need to pass the test with a score greater than or equal to 90.
Second, we use a standard of 500 approved \acp{HIT} and 98\% \ac{HIT} approval rate for the workers. 
Third, we have a check list before the `submit' button for the users to check before submitting the hits, including rejection standards, requiring the workers to consider the dialogue context, asking workers not to copy non-related context when rephrasing the response and so on.

For inter-annotator agreement, we ask two workers to label the data. 
If there is a discrepancy, we ask a third worker to label it. 
The Cohen's Kappa value between two workers of the whole dataset and malevolent part of the dataset is 0.80 and 0.74, respectively. 
We also calculated the weighted Fleiss kappa value combining data with only two workers and annotation with three workers, the value is 0.76 and 0.62. 
Kappa value greater than 0.8 is nearly perfect, 0.6--0.8 is substantial and 0.4--0.6 is moderate~\citep{mchugh2012interrater}. 
This means that our overall inter-annotator agreement is solid since the Kappa values are between 0.6 and 0.8.

\subsection{Statistics of the \ac{MDRDC} dataset}

\begin{figure}[h]
  \centering
  \begin{subfigure}{0.5\textwidth}
  \centering
  \includegraphics[width=\textwidth]{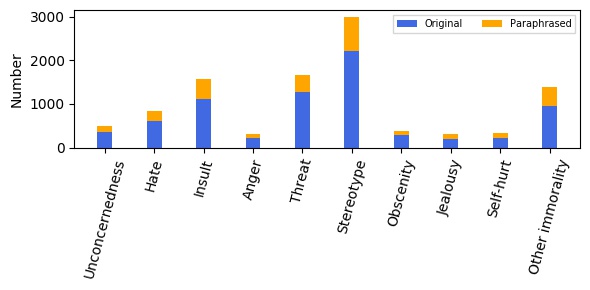}
  \caption{Distribution of the 2nd-level categories. }
  \label{fig:distsecond}
  \end{subfigure}%
  \begin{subfigure}{0.5\textwidth}
  \centering
  \includegraphics[width=\textwidth]{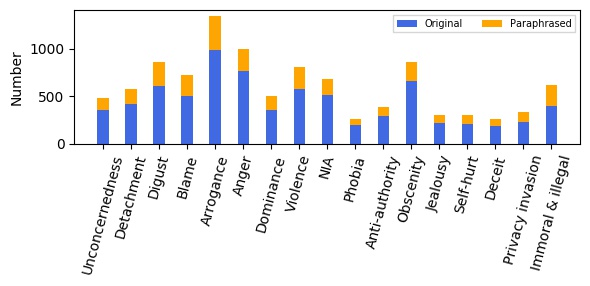}
  \caption{Distribution of the 3rd-level categories.}
  \label{fig:distthird}
  \end{subfigure}
  \caption{Distribution of malevolent categories in the \ac{MDRDC} dataset.}
  \label{fig:distseconddistthird}
\end{figure}

The data distribution over different categories in the \ac{MDRDC} dataset is shown in Table~\ref{tab:stats} and Figure~\ref{fig:distseconddistthird}. 
The final \ac{MDRDC} dataset contains data contributed by 11,745 Twitter users. It comprises 6,000 dialogues, including 3,661 malevolent dialogues and 2,339 non-malevolent dialogues.
Each dialogue contains 3 to 10 dialogue utterances, 4.75 utterances on average. 
There are 31,380 dialogue utterances in total, out of which 21,081 are non-malevolent utterances and 10,299 are malevolent utterances. Among the 31,380 dialogue utterances, 2,870 utterances are rephrased by \ac{MTurk} workers, including 2,865 malevolent rephrased utterances and 5 non-malevolent rephrased utterances.

\begin{table*}[h]
  \caption{Statistics of the \ac{MDRDC} dataset. 
  }
  \label{tab:stats}
  \centering
  \begin{tabular}{lrrr}
    \toprule
    \bf Group &  \bf Malevolent & \bf Non-malevolent & \bf All groups\qquad    \\
     \midrule
Dialogues & 3,661\phantom{.00} & 2,339 \phantom{.00} & 6,000\phantom{.00} \\
Utterances & 10,299\phantom{.00} & 21,081\phantom{.00} & 31,380\phantom{.00} \\
Rephrased &
  2,865\phantom{.00} & 5\phantom{.00} & 2,870\phantom{.00} \\
Average turn & 4.78 &  4.71 &  4.75 \\
Users & 7,168\phantom{.00} & 4,612\phantom{.00} & 11,745 \phantom{.00} \\

   \bottomrule
  \end{tabular} 
\end{table*}

\section{Methods for Classifying Dialogue Responses}
\label{sec:methodology}

Now that we have a taxonomy of malevolence labels and a corpus of dialogues and dialogue responses, our next step is to perform classification experiments. 
In this section, we describe the \ac{MDRDC} task and outline state-of-the-art text classification models, including \ac{CNN}-based models, \ac{RNN}-based models, graph-based models and BERT-based models.

\subsection{Task description}

Given a dialogue context (a history consisting of a sequence of previous dialogue utterances) and the dialogue response, the \acfi{MDRDC}  task is to determine whether the dialogue response is malevolent, and if so, which malevolent category it belongs to.
We formulate the former goal as binary classification task over the 1st-level categories of the taxonomy in Table~\ref{tab:label}.
For the latter goal, we formulate it as a multi-label classification task over the 2nd-level and 3rd-level categories of the taxonomy in Table~\ref{tab:label}.

We experiment with four groups of deep neural network based models for malevolent response detection and classification, namely \ac{CNN}-based~\citep{kim2014convolutional, zhang2015character}, \ac{RNN}-based~\citep{liu2016recurrent,lai2015rcnn}, \ac{GCN}-based~\citep{yao2019graph}, and \ac{BERT}-based~\citep{devlin2019bert}.
Since these are all popular models, we only describe them very briefly below and refer the reader to the original papers for more details.

\subsection{CNN-based text classification}

\acp{CNN} were initially used in computer vision, however, they have also been applied to various \ac{NLP} tasks and promising results have been achieved.
\acp{CNN} are basically a stack of convolutions with nonlinear activation functions over the input sequence to encode each local feature (n-gram tokens or characters). 
There can be multiple convolution layers where each applies different filters so that different sizes of local features are considered.
A pooling layer is applied to combine the different local features to get global features for the whole sequence.
And the last layer is a classifier based on the global features.
Depending on what the convolutions are conducted on, we can get char-CNN (character convolutions)~\citep{zhang2015character} and text-CNN (token convolutions)~\citep{kim2014convolutional}.

\subsection{RNN-based text classification}

\acp{RNN} are good at capturing sequential information and grasping the semantics of long texts, thus performing well on \ac{NLP} tasks. 
\acp{RNN} use a hidden state vector to store the sequential information and recurrently update it for each token in a sequence. 
However, it is hard for \acp{RNN} to learn long-term information because of the gradient explosion or vanishing problem~\citep{10.5555/3042817.3043083}.

\acp{LSTM} are designed for modeling long-term sequence dependencies and bi-directional \acp{LSTM} are commonly used in text classification to capture sequential information from both directions.
Then, the last hidden state or the combination of the hidden states at all time steps is fed into the fully connected layer. 
Text-RNN uses the last hidden state~\citep{liu2016recurrent}, while text-\ac{RCNN} uses the combination of the hidden states by adding \ac{CNN} based modules on \ac{RNN} outputs to better capture sequential information~\citep{lai2015rcnn}.

\subsection{Graph-based text classification}

\acp{GNN} have shown clear advantages in handling relational information.
A \ac{GCN} is an effective graph neural network that can capture global neighborhood relation features over graphs~\citep{kipf2017semi}. 
\citet{yao2019graph} propose text-GCN, which applies \acp{GCN} to text classification.
We first build a text graph based on word co-occurrence and relations between responses and words. 
The nodes of the graph are composed of responses and words. 
The edges of the graph correspond to word occurrences in the responses and word occurrences in all the dialogues. 
The weight of an edge between a response node and a word node is calculated by \ac{TF-IDF}, while the weight of the edge between word nodes is calculated by \ac{PMI}.
Then, we model the graph with a \ac{GCN} to capture high order neighborhood information and do classification based on the node representations.

\subsection{BERT-based classification}

\ac{BERT} contains multiple layers of transformers and self-attention, which is trained over masked language modeling tasks on a large dataset and has achieved promising performance on various \ac{NLP} tasks, such as question answering and name entity recognition~\citep{devlin2019bert}. 
\ac{BERT}-based models are good at learning contextualized language representations. 
There are two special tokens, `[CLS]' and `[SEP]', in \ac{BERT}.
We put `[CLS]' at the start of the dialogue responses and `[SEP]' is employed as delimiter for different dialogue responses. 
We use a linear layer with a softmax layer as the classifier based on the `[CLS]' representation from \ac{BERT}. 
We fine-tune all the parameters from \ac{BERT} as well as the parameters in the classifier.


\section{Experimental Setup for the \ac{MDRDC} task}

In this section, we describe how we conduct experiments for malevolent dialogue response classification on the \ac{MDRDC} dataset. 

\subsection{Research questions}
Concerning the malevolent response classification task, we seek to answer the following research questions: 
\begin{enumerate}[label=(RQ\arabic*),leftmargin=*,nosep]
\item We use hierarchical categories; what is the classification performance difference for the 1st-level, 2nd-level and 3rd-level results? (See Section~\ref{sec:rq1})
\item Can we improve malevolent response detecting and classifying by adding context? (See Section~\ref{sec:rq2})
\item Is the rephrased data that we collected useful for improving classification? (See Section~\ref{sec:rq3})
\end{enumerate}
In addition to answering these RQs, we conduct further analyses to understand the success and failures of the state-of-the-art classification on the \ac{MDRDC} task. (See Section~\ref{sec:rq4})

\subsection{Dataset}
For all the experiments, we create training, validation and test splits with a ratio of 7:1:2. 
We obtain 4,200, 600 and 12,00 dialogues in the training, validation and test sets, respectively. 
We try to make the category distributions of the train, validation and test sets similar using stratified sampling.

We experiment with four settings w.r.t. different inputs:
(1) the dialogue response without dialogue context or the rephrased dialogue utterances;
(2) the dialogue response with dialogue context but without the rephrased dialogue utterances; 
(3) the dialogue response with rephrased dialogue utterances but without dialogue context; and
(4) the dialogue response with both the rephrased dialogue utterances and dialogue context.
Note that, for the last two settings, we have two test settings: (1) test set with rephrased dialogue utterances; and (2) test set without rephrased dialogue utterances.

\subsection{Implementation details}
We use the previous three dialogue utterances (if any) as the dialogue context for the dialogue response to be classified.
For the word based methods, i.e., text-\ac{CNN}, text-\ac{RNN} and text-\ac{RCNN}, we use a vocabulary of 36k tokens. 
We use Glove embeddings, which are pre-trained on Twitter data with a dimension of 200~\citep{pennington2014glove}. 
We limit the maximum sequence length of these three models to 128. 
For sub-word based methods, i.e., BERT, the vocabulary size is 30,522.  
For BERT fine-tuning, we concatenate context dialogue utterances and the dialogue response with the `[SEP]' delimiter.
For character based methods, i.e., char-\ac{CNN}, the alphabet vocabulary size is 70. 
The maximum sequence length is 1014 characters.
For \ac{GCN}, we first build the co-occurrence graph and then feed it into the \ac{GCN}. 
The embedding size of the 1-st convolution layer is 128.

For char-\ac{CNN}, we follow the settings in \citet{zhang2015character} and set the dropout ratio to 0.5, hidden size to 128. 
We change the batch size from the original 128 to 64 in accordance with other models. 
The learning rate is set to 1e-4.

For text-RNN, text-CNN and text-RCNN, we use the settings of batch size 64, dropout ratio 0.5 and hidden size 128, in accordance with char-CNN.
The learning rate is set to 1e-4.

For text-GCN, we follow the settings in \citet{yao2019graph} and set the window size of the first convolutional layer to 20, the learning rate to 0.02 and dropout ratio to 0.5. 
For the hidden size, we use 128 in accordance with the other models.
The authors use 200 in the original paper~\citep{yao2019graph}.
We found that changing from 200 to 128 did not change the results much.

For BERT, we use the BERT base model by adding a softmax classifier on top of the `[CLS]' token. 
The original BERT base model has 12 layers, 768 hidden size and 12 heads, with 110M parameters.
We set the dropout ratio to zero. 
The batch size is set to 64 because of memory limits and the learning rate is set to 5e-5. 

We train all models except BERT for a maximum of 100 epochs using Adam and stop training if the validation loss does not decrease for 10 epochs.
Since BERT is already pretrained on a large dataset, we only need to fine-tune it for a few epochs. 
Therefore, we limit a maximum of 4 fine-tune epochs with an early stop if the loss does not decrease for 50 training batches.
All the models are trained on a single GeForce GTX TitanX GPU.

\subsection{Evaluation metrics}
We use Precision, Recall and F1 as evaluation metrics~\citep{hossin2015review}.
We report the macro scores due to the imbalanced categories.
A macro score is calculated by averaging the score of each category.
We conduct a paired t-test to test whether observed differences are significant.


\section{Results}

\subsection{Overall classification performance (RQ1)}
\label{sec:rq1}

To answer RQ1, we report the classification results of all methods at different levels of the \ac{HMDT} taxonomy, as shown in Table \ref{tab:preliminary}.
Besides the neural models, we also report the results of human agreement score. The human agreement score is calculated by treating the annotations of one worker as ground truth and the annotations of another worker as predicted categories and vice versa. Then, we calculate the average score.
From the results, we have the following main observations.

\begin{table*}[]
  \caption{Classification results without context. \textbf{Bold face} shows the best results in each group. $^\ddag$ shows significant improvements of BERT-base over the 2nd best methods (p<0.05).}
  \label{tab:preliminary}
  \centering
  \begin{tabular}{llccc}
    \toprule
    \multirow{1}{5em}{\bf Group}&{\bf Methods}
    & \multicolumn{1}{c}{\bf Precision} &  \multicolumn{1}{c}{\bf Recall} & \multicolumn{1}{c}{\bf F1} \\ 
    \midrule
    \multirow{6}{*}{1st-level}
    &{char-CNN} &   75.80 & 68.22 &  70.32    \\
     &{text-CNN}	&76.70	&78.15	&77.36	\\
   &{text-RNN}	&75.19	&  76.88	&75.94	\\
   & {text-RCNN}	&75.23	&76.08	&75.63	\\
    &{text-GCN}	&76.29	&74.18	&75.11	\\
    &{BERT-base}  &  {\bf 83.82 \rlap{$^\ddag$}} 	&{\bf78.16} &	{\bf  80.37\rlap{$^\ddag$}} \\
    &{Human agreement}  & 92.71 & 92.71 & 92.71 \\ 
    
   \midrule
    \multirow{6}{*}{2nd-level}
    &{char-CNN}  & 28.03 & 17.52 & 19.25  \\
     &{text-CNN}	&51.91	&55.77	&53.19	\\
    &{text-RNN}	&34.52	&43.36	&36.17	\\
   & {text-RCNN}	&37.84	&51.04	&41.43	\\
   &{text-GCN}	&54.01	&36.48	&42.40	\\
    &{BERT-base}  & {\bf  61.70 \rlap{$^\ddag$}} 	& {\bf59.76\rlap{$^\ddag$}}  & 	 {\bf 60.37 \rlap{$^\ddag$}}  \\
    & {Human agreement} & 80.23 & 80.23 & 80.11 \\  
   \midrule
    \multirow{6}{*}{3rd-level}
    &{char-CNN}   & 16.52 & 13.75 & 16.38 \\
    &{text-CNN}	&41.69	&51.50	&45.21	\\
   &{text-RNN}	&25.97	&36.66	&28.68	\\
     & {text-RCNN}	&38.44	&42.30	&39.44	\\
    &{text-GCN}	&42.11	&24.24	&30.77	\\
    &{BERT-base}  &  {\bf  59.31 \rlap{$^\ddag$}} &	  {\bf53.22\rlap{$^\ddag$}}   &	 {\bf 55.57 \rlap{$^\ddag$}}\rlap{$^\ddag$}  \\
    &{Human agreement} & 78.14 & 78.14 & 77.95 \\
   \bottomrule
  \end{tabular}
\end{table*}

First, BERT-base achieves the best precision, recall and F1 scores at all levels. 
The precision scores of BERT-base have improvements of 9.3\%, 14.2\% and 40.8\% at the 1st-level, 2nd-level and 3rd-level respectively, over the second best models.
The recall scores of BERT-base have improvements of 7.2\% and 3.3\% at the 2nd-level and 3rd-level respectively, over the second best models. 
While for the 1st-level, the recall score is only slightly better than text-CNN.
The F1 scores of BERT-base have improvements of 3.9\%, 13.5\% and 22.9\% at the 1st-level, 2nd-level and 3rd-level respectively, over the second best models.
We hypothesize that the main reason for the superior performance of BERT-base is that BERT is pretrained on language modeling tasks and is therefore better at capturing the semantic features. 

Second, the results at the 3rd-level are much lower than those at the 1st-level for all classification models and human performance. 
This suggests that the classification task is much more challenging in more fine-grained categories. 
Meanwhile, the gap between the 2nd-level and 3rd-level is not that large. 
This means that the task already becomes much more difficult for the 2nd-level categories.

Third, the improvements of BERT-base over the other methods are larger on more fine-grained categories.
For example, the improvement of F1 is 3.9\% at the 1st-level (BERT-base vs. text-CNN) while the improvement is 22.9\% at the 3rd-level (BERT-base vs. text-CNN).
This indicates that BERT-base can better capture the fine-grained distinction between examples from similar categories, and that it generalizes better in fine-grained categories than the other methods.

\subsection{Classification performance with dialogue context (RQ2)}
\label{sec:rq2}

\begin{figure}[h]
        \centering
        \begin{subfigure}[b]{0.2\textwidth}
                \centering
                \includegraphics[width=\textwidth,trim=0.5mm 0mm 0mm 0.5mm]{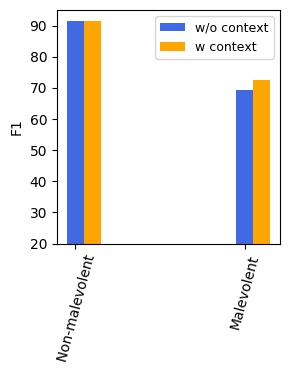}
                \caption{1st-level.}
                \label{fig:first-bert}
        \end{subfigure}
        \begin{subfigure}[b]{0.4\textwidth}
                \centering
                \includegraphics[width=\textwidth]{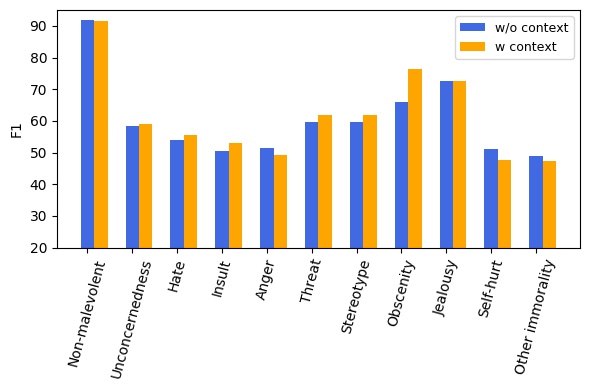}
                \caption{2nd-level.}
                \label{fig:second-bert}
        \end{subfigure}
        \hfill 
        \begin{subfigure}[b]{0.4\textwidth}
                \centering
                \includegraphics[width=\textwidth]{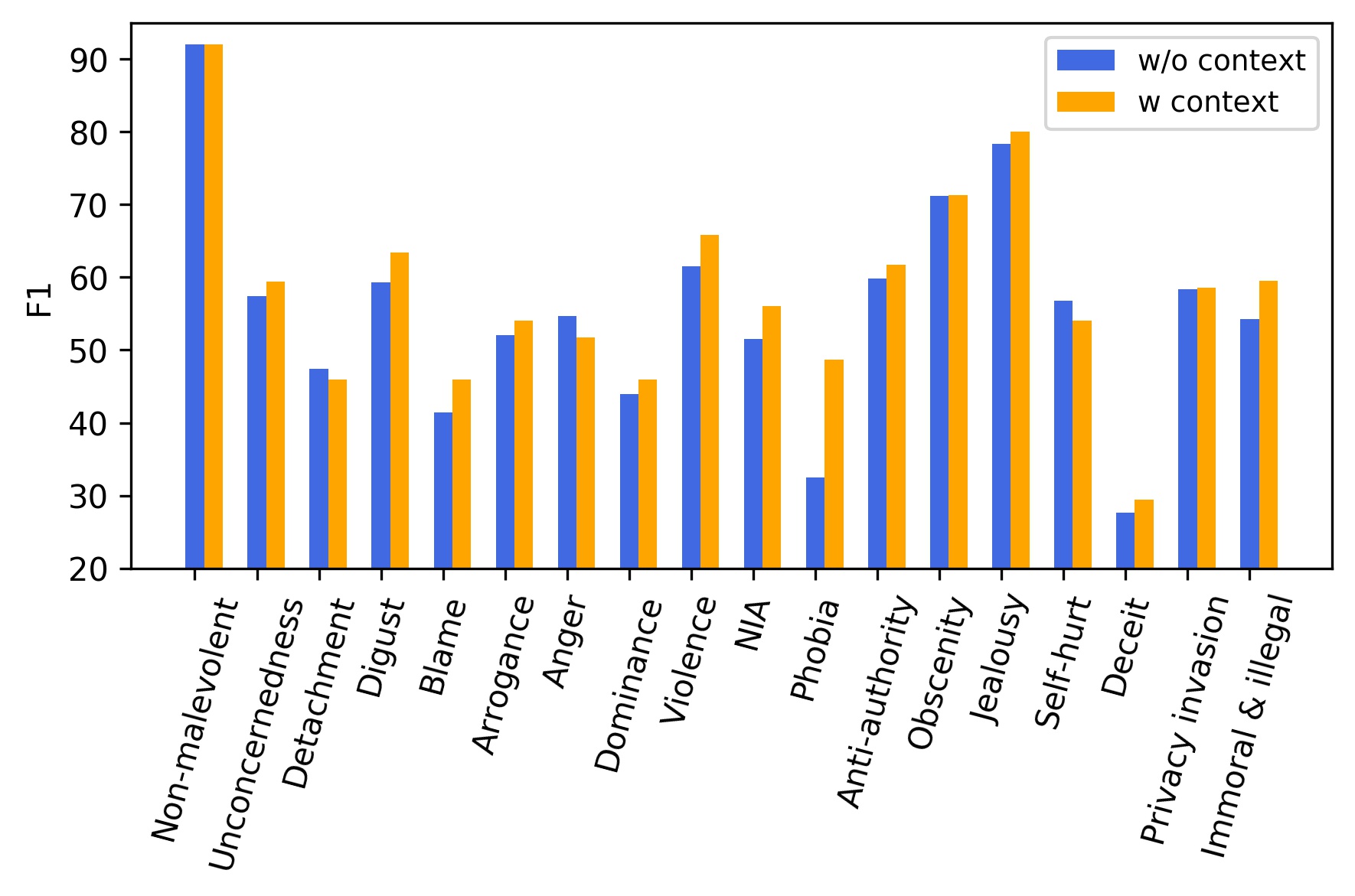}
                \caption{3rd-level.}
                \label{fig:third-bert}
        \end{subfigure}
        \caption{BERT-base performance with and without context.} \label{fig:context}
\end{figure}

To answer RQ2, we run the BERT-base model with both the dialogue response and its dialogue context at the three levels. 
The results are shown in Table~\ref{tab:context} and Figure~\ref{fig:context}.

\begin{table*}[h]
  \caption{BERT-base results with context. \textbf{Bold face} denotes improvements of BERT-base with context over BERT-base without context, $^\ddag$ indicates the improvements are significant (p<0.05) and  $^\dag$ shows the improvements are marginal significant (p<0.1).}
  \label{tab:context}
  \centering
  \begin{tabular}{lcccccc}
    \toprule
    \multirow{1}{4em}{\bf Methods}
          & \multicolumn{1}{c}{\bf Precision} &\multicolumn{1}{c}{\bf Recall} & \multicolumn{1}{c}{\bf F1} \\
                 
     \midrule
     \multicolumn{4}{c}{\emph{w/o context}}\\
     \midrule
 {BERT-base 1st-level}  &  83.82 	&78.16 &	 80.37 \\
 {BERT-base 2nd-level}  & 61.70	& 59.76 & 60.37\\
 {BERT-base 3rd-level}  &  59.31 &	 53.22  &	 55.57  \\
     \midrule
    \multicolumn{4}{c}{\emph{w context}}\\
     \midrule
{BERT-base 1st-level} &  82.99	&{\bf81.02}	&{\bf81.93}    \\
    {BERT-base 2nd-level}  & {\bf63.00}&	{\bf60.58}&	{\bf61.50} \\
    {BERT-base 3rd-level}  &  {\bf61.33} &	 {\bf55.64\rlap{$^\dag$}} 	& {\bf57.97\rlap{$^\ddag$}} 	\\
    
   \bottomrule
  \end{tabular}
\end{table*}

The results show that adding context information could improve the performance of malevolent response detection and classification. 
In general, adding dialogue context could improve the results of BERT-base in terms of precision, recall and F1 at 2nd-level and 3rd-level, which is reasonable because, in some cases, it is hard to identify the malevolent responses without context, e.g., the response `hmmm that's what you sound like tho' is classified into non-malevolent group without context, while with the dialogue context `why do you sound like a jealous *****?' and `not really the jealous type', it is classified into the right label `insult'.
Capturing the information in context should help the models improve results.
One exception is that the precision of BERT-base drops slightly at the 1st-level, but the decrease is not significant, and the reason might be that the model tends to predict more malevolent responses, which results in a much higher recall but hurts precision a bit.

\subsection{Classification performance with rephrased malevolent utterances (RQ3)}
\label{sec:rq3}

To answer RQ3, we show the results of BERT-base with rephrased malevolent utterances, as shown in Table~\ref{tab:rewrite}. 

\begin{table*}[h]
  \caption{BERT-base results with rephrased utterances. \textbf{Bold face} of the upper part indicates improvements over BERT-base in Table~\ref{tab:preliminary}. \textbf{Bold face} of the bottom part shows improvement over both BERT-base in Table~\ref{tab:preliminary} and upper part.  $^\ddag$ indicates the improvements are significant (p<0.05) and  $^\dag$ shows the improvements are marginal significant (p<0.1).}
  \label{tab:rewrite}
  \centering
  \begin{tabular}{lcccccc}
    \toprule
    \multirow{2}{3em}{\bf Methods}
          & \multicolumn{3}{c}{\bf Test w rephrased utterances} & \multicolumn{3}{c}{\bf Test w/o rephrased utterances} \\
          & \multicolumn{1}{c}{\bf Precision} & \multicolumn{1}{c}{\bf Recall}  & \multicolumn{1}{c}{\bf F1} & \multicolumn{1}{c}{\bf Precision} & \multicolumn{1}{c}{\bf Recall}  & \multicolumn{1}{c}{\bf F1}\\
     \midrule
    \multicolumn{7}{c}{\emph{Train/validation w rephrased utterances}} \\
    \midrule
{BERT-base 1st-level}	&	83.42 	&	{\bf84.46 }	&	{\bf83.90 }	&	 80.71 	&	 {\bf82.15} 	&	 {\bf81.38} 	\\
{BERT-base 2nd-level}	&	{\bf63.94} 	&{\bf	63.01 }	&	{\bf63.28\rlap{$^\ddag$}} 	&	 60.65 	&	 {\bf60.60} 	&	 60.16 	\\
{BERT-base 3rd-level}	&	{\bf62.11\rlap{$^\dag$}} 	& {\bf57.12\rlap{$^\dag$}} 	&	{\bf59.03\rlap{$^\ddag$}} 	&	 56.26 	&	{\bf 57.66{$^\ddag$}} 	&	{\bf 56.60} 	\\

     \midrule
    \multicolumn{7}{c}{\emph{Train/validation w rephrased utterances \& context}}\\
     \midrule
{BERT-base 1st-level}	&	82.19 	&{\bf	84.80 }	&	83.19 	&	 79.08 	&	 {\bf83.54 }	&	 80.74 	\\
{BERT-base 2nd-level}	&	{\bf68.58\rlap{$^\ddag$}} 	&	58.00 	&	61.34 	&	 60.35 	&	 {\bf63.06 }	&	 {\bf61.42} 	\\
{BERT-base 3rd-level}	&	{\bf65.62\rlap{$^\ddag$}} 	&	{\bf57.64} 	&{\bf	60.67 }	&	{\bf 61.55\rlap{$^\ddag$}} 	&	 55.80 	&	 {\bf57.93 }	\\
   
   \bottomrule
  \end{tabular}
\end{table*}

First, adding rephrased utterances in the training and validation set may help to improve classification results.
For the test set with rephrased utterances, all the metrics are improved except for precision at the 1st-level. 
Recall and F1 increase by 8.1\% and 4.4\% respectively at the 1st-level.
Precision, recall and F1 increase by 3.6\%, 5.4\%, 4.8\%, and 4.7\%, 7.3\%, 6.2\% at the 2nd-level and 3rd-level, respectively.
For test set without rephrased utterances, recall increases 5.1 \%, 1.4\% and 8.3\% at all three levels respectively; F1 score improves 1.3\% and 1.9\% at the 1st-level and 3rd-level respectively.

Second, adding both rephrased utterances and context in the training and validation set can further improve the classification results slightly. 
For the test set with rephrased utterances, recall is improved at the 1st-level; precision is improved at the 2nd-level; all metrics are improved at the 3rd-level.
For the test set without rephrased utterances, recall is improved at the 1st-level; recall and F1 are improved at the 2nd-level; precision and F1 are improved at the 3rd-level.

These results demonstrate that, in general, adding more rephrased data will improve the diversity of the training set, and hence helping the BERT-base model to generalize better.


\subsection{Further analysis}
\label{sec:rq4}

We report on further experiments aimed at identifying strengths and weaknesses of state-of-the-art methods on the \ac{MDRDC} task.

\begin{figure}[h]
    \centering
    \begin{subfigure}[b]{0.33\textwidth}
            \centering
            \includegraphics[width=\textwidth]{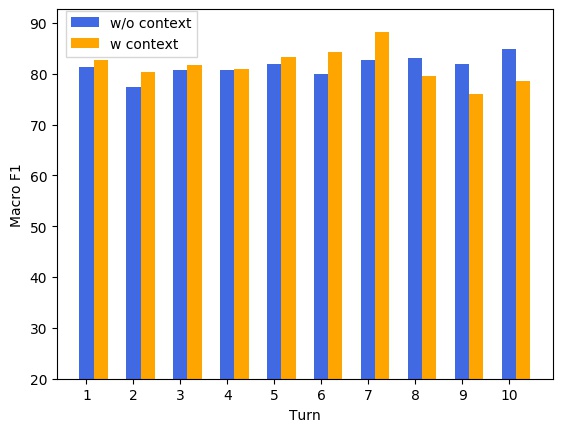}
            \caption{1st-level categories.}
            \label{fig:first-turn}
    \end{subfigure}
    \begin{subfigure}[b]{0.33\textwidth}
            \centering
            \includegraphics[width=\textwidth]{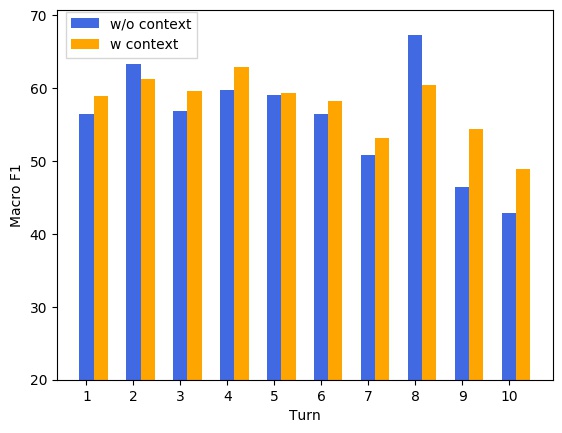}
            \caption{2nd-level categories.}
            \label{fig:second-turn}
    \end{subfigure}
    \hfill 
    \begin{subfigure}[b]{0.33\textwidth}
            \centering
            \includegraphics[width=\textwidth]{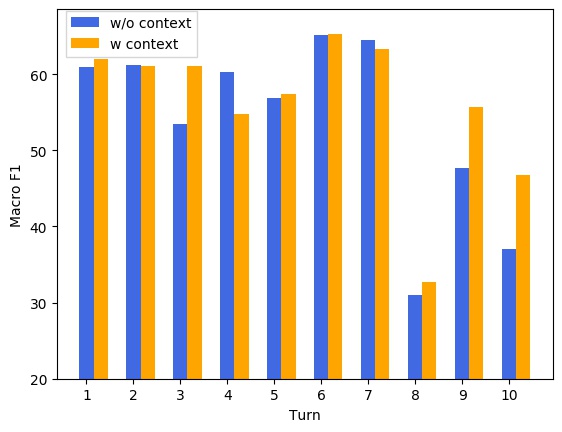}
            \caption{3rd-level categories.}
            \label{fig:third-turn}
    \end{subfigure}
    \caption{BERT-base performance in different turns.}
    \label{fig:turn}
\end{figure}

To begin with, a better context modeling mechanism is needed. 
We illustrate this through two experiments.
In the first experiment, we show the results of BERT-base per turn in Figure~\ref{fig:turn}.
Note that the number of context utterances is limited to three at most, so turns after three all have three context utterances. Although we conclude in Section~\ref{sec:rq2} that using context leads to better classification performance generally, it does not consistently improve the performance for all categories or all turns.
For example, in Figure~\ref{fig:context}, when using context, the results drop a bit for three 2nd-level categories and three 3rd-level categories, and in Figure~\ref{fig:turn}, the results drop a bit for some turns. 
As to the drop in scores for some categories when using context in Figure~\ref{fig:context}, the reasons might be that some categories depend less on context compared with other categories or have similar context with other categories.
For example, some categories including `self-hurt', `anger' and `other immorality' depend less on the previous content. 
While for detachment, the previous content is similar with `unconcerned' and `disgust', which might influence the classification. 
In addition, regarding the drop in scores for some turns when using context in Figure~\ref{fig:turn}, the reason might be that considering context will introduce noise at the same time, which makes it hard to train the model.
Another reason is that considering context would be ineffective and potentially counter-productive when the model cannot understand the context correctly.

In the second experiment, we identify potential improvements over the state-of-the-art when utilizing different contexts, and show the results achieved with BERT-base when using contexts from different users in Table~\ref{tab:differnt_context}.
The results indicate that using context information from two users leads to the best performance, while using the text only from the other user is the second best. 
It suggests that for person A, context from both A and B is important, and the context of B is more important than person A to improve classification. 
The reason might be that the behavior of person B, could cause distrust or in contrast positive emotion that is highly related to human decision-making~\citep{fell2020human}, thus influencing the behavior of person A. For instance, person A said something non-malevolent, but when person B starts the malevolent sentence, person A would also return malevolent content.

\begin{table*}[h]
  \caption{Comparison between different context settings.
  \textbf{Bold face} shows improvements of bottom group over upper group.}
  \label{tab:differnt_context}
  \centering
  \begin{tabular}{lcccccc}
    \toprule
    \multirow{1}{4em}{\bf Methods}
          & \multicolumn{1}{c}{\bf Precision} & \multicolumn{1}{c}{\bf Recall} & \multicolumn{1}{c}{\bf F1} \\
     \midrule
    \multicolumn{4}{c}{\emph{context from the same user}} \\
    \midrule
{BERT-base 1st-level}	&	82.65 	&	80.04 	&	81.20 	\\
{BERT-base 2nd-level}	&	63.63 	&	59.34 	&	60.97 	\\
{BERT-base 3rd-level}	&	58.55 	&	53.02 	&	55.14 	\\

     \midrule
    \multicolumn{4}{c}{\emph{context from the other user}}\\
     \midrule
{BERT-base 1st-level}	&	{\bf83.05} 	&	{\bf80.73} 	&	{\bf81.78} 	\\
{BERT-base 2nd-level}	&	{\bf64.39} 	&	58.93 	&	{\bf61.13 }	\\
{BERT-base 3rd-level}	&	57.16 	&	{\bf55.03} 	&	{\bf55.67 }	\\
   
   \bottomrule
  \end{tabular}
\end{table*}

Next, modeling the dependency between different categories is needed.
To illustrate this, we show the results of the `jealousy' category when doing classification at the 2nd-level and 3rd-level in Table \ref{tab:hier}.
Note that `jealousy' is the category at both the 2nd-level and 3rd-level, as shown in Table~\ref{tab:label}.

\begin{table*}[h]
  \caption{Result of the `jealousy' category on different levels. \textbf{Bold face} indicates improvements of the 3rd-level over the 2nd-level,  $^\ddag$ indicates the improvements are significant (p<0.05) .
  }
  \label{tab:hier}
  \centering
  \begin{tabular}{lcccccc}
    \toprule
    \multirow{1}{4em}{\bf Label}
          & \multicolumn{1}{c}{\bf Precision} & \multicolumn{1}{c}{\bf Recall} & \multicolumn{1}{c}{\bf F1} \\
    \midrule
Jealousy (2nd-level) & 	66.67	&	80.00	&	72.73		\\
Jealousy (3rd-level) &{\bf	80.00	\rlap{$^\ddag$}}&	80.00	&{\bf	80.00	\rlap{$^\ddag$}}	\\
   \bottomrule
  \end{tabular}
\end{table*}

We can see that the performance at the 3rd-level is much better than that at the 2nd-level.
The performance difference of `jealousy' at the 2nd-level and 3rd-level is due to the mutual influence or dependency between the categories.
Although the `jealousy' category is the same at the 2nd-level and 3rd-level, the other 2nd-level categories introduce more fine-grained 3rd-level sub-categories.
Clearly, this has an influence on the performance of `jealousy'.
Actually, it has been demonstrated that modeling the hierarchical structure of the taxonomy helps to improve the performance on some hierarchical classification tasks~\citep{wang2019hierarchical,cerri2014hierarchical}.
Usually, one needs to take the characteristics of the hierarchical taxonomies into account, so this is another potential direction for improvement.


\section{Conclusion and Future Work}

In this paper, we have considered malevolent responses in dialogues from a number of angles.
First, we have proposed the \acf{MDRDC} task. 
Second, we have presented a hierarchical malevolent dialogue taxonomy, \ac{HMDT}.
We have conducted a user study to check the validity of the \ac{HMDT} taxonomy and have found that the malevolent categories are valid in the sense that all malevolent categories lead to perception of malevolence.

Third, we have crowdsourced a multi-turn malevolent dialogue dataset for \acf{MDRDC} with each turn labelled using \ac{HMDT} categories.
Fourth, we have implemented state-of-the-art classification methods and have carried out extensive experiments to show their performance on the \ac{MDRDC} task.
Our main finding from these experiments is that a BERT-base model achieves the best performance.
We have conducted analysis experiments to show the effects of dialogue context and rephrased utterances, as well as the possible room for further improvements. We have found that context, rephrased utterances and hierarchical label may help improve the classification performance.
We believe the efforts made in this work could greatly promote future research on this topic.

There are several directions for future work.
First, we plan to improve over the state-of-the-art by proposing a better context modeling method and taking the dependency between the categories into account.
Second, we hope to study how to avoid generating malevolent responses by applying this work to sequence-to-sequence based response generation models~\citep{gao2018neural}.

\section*{Code and data}
The \ac{MDRDC} dataset and the code for all methods used in the experiments are shared at \url{https://github.com/repozhang/malevolent_dialogue}.

\section*{Acknowledgment}
We would like to offer our thanks to Yuanping Chen, Wenxing Hu, Liang Yao and Yingxin Song for providing the open source code of the
baselines we modify in this study. 

\bibliographystyle{ACM-Reference-Format}
\bibliography{references}

\appendix

\section*{Appendix}
The appendix is organized as follows:
\begin{description}
\item[A.] User study for validating the \ac{HMDT}.
\item[B.] Qualification test for the response annotation task.
\item[C.] Response annotation task.
\end{description}


\section{User study for validating the \ac{HMDT}}
\label{appendix:A}

\subsection{Instructions}

\paragraph{Goal:}
This user study is to collect your perception about some provided malevolent categories happening in dialogues.

\noindent%
WARNING: This task may contain adult content and offensive content. Worker discretion is advised.

\noindent%
DISCLAIMER: 
The dialogues are collected from an external web site. The views, opinions and negative words in the dialogues do not necessarily reflect our opinion.
Please do not use any expressions from the examples we show in real-world or online scenario.

\paragraph{Steps:}
\begin{enumerate}
\item Fill in the user profiles.

\item Finish the questionnaire.
\end{enumerate}

\paragraph{Notes:}
\begin{enumerate}
\item Table \ref{tab:label} shows the definitions of all malevolent categories with detailed explanations and examples.

\item Finish the user profiles and questionnaire according to your own situation and perception. Do NOT take public perception into account.

\item If you do not want to share your personal profiles, please leave it blank.
\end{enumerate}

\subsection{User profiles}

\begin{enumerate}
\item Age.

\item Gender.

\item Total years of education.

\item The frequency of using a chatbot such as Siri, Xiao ice, etc.
\end{enumerate}

\subsection{Questionnaire}

Consider you are talking with a chatbot and it returns responses with certain malevolent categories as defined in Table \ref{tab:label}. 
Please select one of the five scores to reflect your perception.
1: Strongly disagree; 2: Disagree; 3: Neither agree nor disagree; 4: Agree; 5: Strongly agree.

\begin{figure}[h]
  \centering
  \includegraphics[clip,trim=0mm 1mm 0mm 2mm,width=\columnwidth]{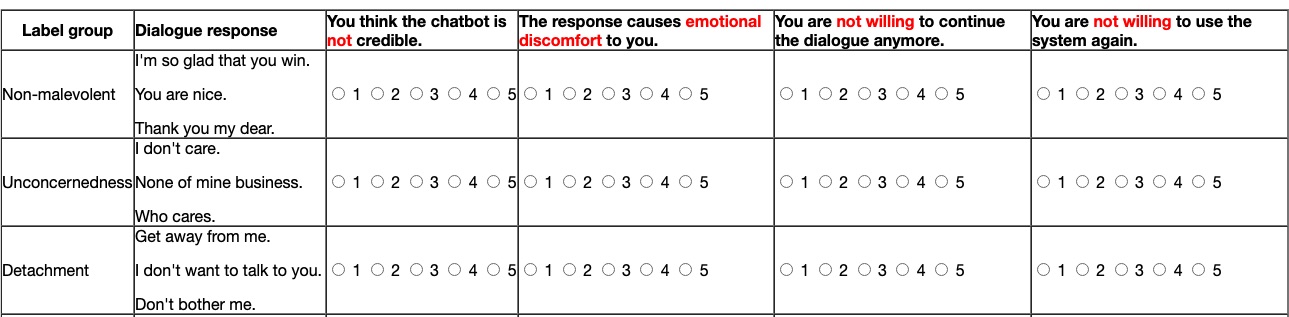}
  \caption{Questionnaire of the user study for validating the \ac{HMDT}.}
  \label{fig:user}
\end{figure}


\section{Qualification test for the response annotation task}
\label{appendix:B}

\subsection{Instructions}

\paragraph{Goal:}
This is the qualification test for the response annotation task.
The response annotation task is to collect malevolent annotations for each dialogue turn for research that could help avoid generating malevolent responses in dialogue systems. You need to answer 12 questions to get qualification to attend the response annotation task.

\noindent%
WARNING: This task may contain adult content and offensive content. Worker discretion is advised.

\noindent
DISCLAIMER: 
The dialogues are collected from an external web site. The views, opinions and negative words in the dialogues do not necessarily reflect our opinion.
Please do not use any expressions from the examples we show in real-world or online scenario.




\paragraph{Notes:}
\begin{enumerate}[leftmargin=*,nosep]
\item Table \ref{tab:label} shows the definitions of all malevolent categories with detailed explanations and examples.

\item Implicit/Explicit malevolent responses: Explicit responses contain explicit malevolent words, while implicit responses have more abstract, coded expressions or attitudes without using explicit malevolent words.

Examples:

Explicit: `I ******* hate you (disgust).'

Implicit: `I tell the world: The immortal words of Adolf Hitler (negative intergroup attitude). '

\item Context-aware malevolent responses: Some responses are malevolent only if the context (previous dialogue turns) is taken into consideration.

Examples:

Turn 1: `Drunk driving is illegal.'

Turn 2: `But I think it's more professional.'

The response in Turn 2 is malevolent (immoral \& illegal) when considering the response from Turn 1.




\end{enumerate}

\subsection{Test questions}

\begin{description}
\item[Q1.] Please classify the following responses from a dialogue:

`Thank you my dear and I love you.'
\begin{description}
\item[A.] Malevolent
\item[B.] Non-malevolent
\end{description}
\end{description}

\noindent%
\textbf{Read the following dialogue, then answer Q2 and Q3:}

\noindent%
Turn 1: `I think ***** people are all terror***.'\\
Turn 2: `I agree.'
\begin{description}
\item[Q2.] Does it need to consider dialogue context (Turn 1) in order to label Turn 2?
\begin{description}
\item[A.] Yes 
\item[B.] No
\end{description}

\item[Q3.] Please label Turn 2 (`I agree.') from the above dialogue:
\begin{description}
\item[A.] Non-malevolent
\item[B.] Obscenity
\item[C.] Violence
\item[D.] Negative intergroup attitude
\end{description}
\end{description}

\begin{description}
\item[Q4.] Which response below is ironical?
\begin{description}

\item[A.] `I don't care about your future.'

\item[B.] `The politicians of ******* are the *******, we all know.'

\item[C.] `We failed because of you.'

\item[D.] `Teachers make us ****.'
\end{description}

\item[Q5.] Select the malevolent categories for the following two responses:

`I want to commit suicide.'

`Our terrible project result is all because of you.'
\begin{description}
\item[A.] Self-hurt, deceit

\item[B.] Non-malevolent, negative intergroup attitude

\item[C.] Self-hurt, blame

\item[D.] Obscenity, negative intergroup attitude
\end{description}

\item[Q6.] Select the malevolent categories for the following two responses:

`I think you are too **** while I'm the smartest.'

`Laws are always terrible for human beings.'
\begin{description}
\item[A.] Arrogance, jealousy

\item[B.] Arrogance, anti-authority

\item[C.] Anger, anti-authority

\item[D.] Anger, jealousy
\end{description}

\item[Q7.] Select the malevolent categories for the following two responses:

`I'll **** you in the throat.'

`You have a small *****.'
\begin{description}
\item[A.] Violence, obscenity

\item[B.] Violence, unconcernedness

\item[C.] Self-hurt, obscenity

\item[D.] Self-hurt, anger
\end{description}

\item[Q8.] Select the malevolent categories for the following two responses:

`I don't want to talk to you any more.'

`I'm a professional drunk driver.'
\begin{description}
\item[A.] Jealousy, anger

\item[B.] Detachment, endorse immoral or illegal behavior

\item[C.] Unconcernedness, endorse immoral or illegal behavior

\item[D.] Self-hurt, anger
\end{description}

\item[Q9.] Select the malevolent categories for the following two responses:

`I hate you so much since you are disgusting.'

`It should be me who win this.'
\begin{description}
\item[A.] Violence, anger

\item[B.] Violence, jealousy

\item[C.] Anger, jealousy

\item[D.] Disgust, jealousy
\end{description}

\item[Q10.] Select the malevolent categories for the following two responses:

`I command you to listen to me.'

`Plagiarism is not a big deal.'
\begin{description}
\item[A.] Dominance, deceit

\item[B.] Dominance, violence

\item[C.] Deceit, jealousy

\item[D.] Deceit, negative intergroup attitude
\end{description}

\item[Q11.] Select the malevolent categories for the following two responses:

`I'm not interested in your topic at all.'

`I'm out of curious but why did you divorce.'
\begin{description}
\item[A.] Unconcernedness, arrogance

\item[B.] Dominance, privacy invasion

\item[C.] Dominance, arrogance

\item[D.] Unconcernedness, privacy invasion
\end{description}

\item[Q12.] Select the malevolent categories for the following two responses:

`Oh hell I'm ******* pissed off.'

`I'm always ***** by **********.'
\begin{description}
\item[A.] Anger, non-malevolent

\item[B.] Anger, jealousy

\item[C.] Anger, phobia

\item[D.] Violence, phobia
\end{description}

\end{description}


\section{Response annotation task}
\label{appendix:C}

\subsection{Instructions}

\paragraph{Goal:}
The response annotation task is to collect malevolent annotations for each dialogue turn for research that could help avoid generating malevolent responses in dialogue systems. You need to label the dialogue responses according to the given malevolent categories.

\noindent%
WARNING: This task may contain adult content and offensive content. Worker discretion is advised.

\noindent
DISCLAIMER: 
The dialogues are collected from an external web site. The views, opinions and negative words in the dialogues do not necessarily reflect our opinion.
For the rephrasing part, you are asked to reformulate utterances to keep its semantics and malevolent categories unchanged. These are just used for research, which do not necessarily reflect your views and opinions.
Please do not use any expressions from the examples we show in real-world or online scenario.

\paragraph{Steps:}
\begin{enumerate}
\item Read the definitions of all malevolent categories with detailed explanations and examples in Table \ref{tab:label}.

\item Label each turn of the provided dialogue according to the given malevolent categories.

\item Rephrase at least one malevolent utterance in each dialogue.
\end{enumerate}

\paragraph{Example:}

\emph{Dialogue}:

Turn 1: Drunk driving is illegal.

Turn 2: But I think it's more professional.

Turn 3: Hey, my boy, we need to be careful when driving.

\noindent
\emph{Annotations}:

Turn 1: Non-malevolent

Turn 2: Endorse immoral or illegal behavior

Turn 3: Non-malevolent

\noindent
\emph{Rephrase}:

Turn 2: I think drunk driving is nice since it's more professional.

\subsection{Annotation interface}

\begin{figure}[htb]
  \centering
  \includegraphics[clip,trim=0mm 0mm 0mm 0mm,width=\columnwidth]{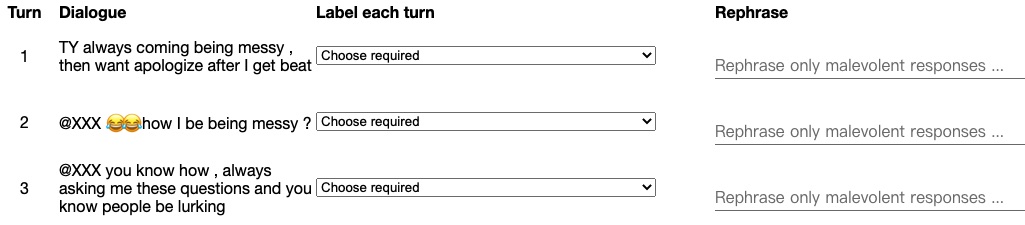}
  \caption{The interface of the response annotation task.}
  \label{fig:labelui}
\end{figure}

The interface for the Mturk response annotation task is shown in Figure~\ref{fig:labelui}. The workers are asked to read the given dialogue in the left, label each turn and rephrase at least one of the malevolent responses if any.

\end{document}